\def\year{2021}\relax
\documentclass[letterpaper]{article} 
\usepackage{aaai21}  
\usepackage{times}  
\usepackage{helvet} 
\usepackage{courier}  
\usepackage[hyphens]{url}  
\usepackage{graphicx} 
\usepackage{natbib}  
\usepackage{caption} 
\title{EasyASR: A Distributed Machine Learning Platform for End-to-end Automatic Speech Recognition}
\author {Chengyu Wang,\textsuperscript{\rm 1} Mengli Cheng,\textsuperscript{\rm 1} Xu Hu,\textsuperscript{\rm 2}\thanks{The work was conducted when X. Hu was afflicated with Alibaba Group.} Jun Huang\textsuperscript{\rm 1}\thanks{J. Huang is the corresponding author.}\\}
\affiliations {
    \textsuperscript{\rm 1} Alibaba Group   \textsuperscript{\rm 2} ByteDance Inc.\\
    \{chengyu.wcy, mengli.cml\}@alibaba-inc.com, huxu.hx@bytedance.com, huangjun.hj@alibaba-inc.com
}
\begin{document}

\maketitle

\begin{abstract}
We present EasyASR, a distributed machine learning platform for training and serving large-scale Automatic Speech Recognition (ASR) models, as well as collecting and processing audio data at scale. Our platform is built upon the Machine Learning Platform for AI of Alibaba Cloud. Its main functionality is to support efficient learning and inference for end-to-end ASR models on distributed GPU clusters. It allows users to learn ASR models with either pre-defined or user-customized network architectures via simple user interface. On EasyASR, we have produced state-of-the-art results over several public datasets for Mandarin speech recognition.
\end{abstract}

\section{Introduction}

As a fundamental task in speech and language processing, Automatic Speech Recognition (ASR) aims to generate transcripts from human speech. Recently, the successful application of deep neural networks has pushed the accuracy of end-to-end ASR models to a new level, but brings significant challenges for building large-scale, robust ASR systems, especially for industrial applications. Major bottlenecks are twofold: i) abundant labeled training data for learning large, accurate ASR models; and ii) an efficient distributed, computing framework for model training and serving at scale.

In this demo, we present EasyASR, a distributed machine learning platform to address both challenges. EasyASR is built upon the Machine Learning Platform for AI (PAI) of Alibaba Cloud~\footnote{https://www.alibabacloud.com/product/machine-learning/}, which provides an ultra-scale, deep learning framework on distributed GPU clusters. Our platform supports the complete process of training, evaluating and serving ASR models. Additionally, it is integrated with the functionalities i) to extract high-quality audio aligned with transcripts from massive video data and ii) to expand existing ASR training sets with various augmentation policies. We have designed easy-to-use PAI components that enable users to build or run ASR models within only a few lines of command, which 
hides complicated techniques from starters. We also provide add-on configurations with the PAI commands to allow advanced users to customize network architectures for their own models. On EasyASR, we achieve state-of-the-art performance for Mandarin speech recognition over multiple public datasets.

%

\section{Platform Description}
In this section, we introduce the EasyASR platform in detail.

\begin{figure}
	\centering
	\includegraphics[width=\columnwidth]{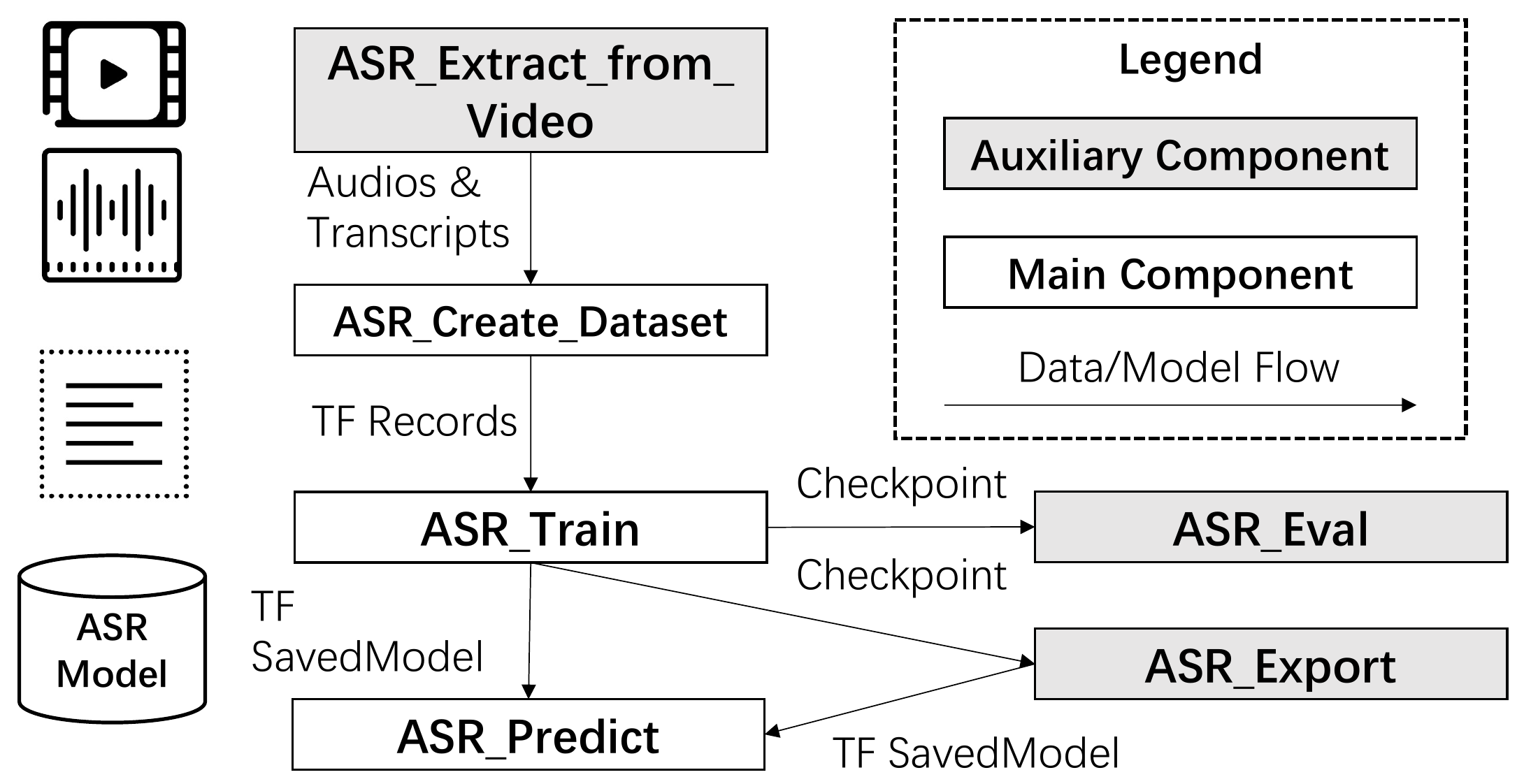}
 	\caption{An overview of PAI's components in EasyASR. }
	\label{fig:framework}
\end{figure}

\noindent\textbf{Function Design.}
On the EasyASR platform, each module is encapsulated as a PAI component, with the overall framework illustrated in Figure~\ref{fig:framework}. Among the three main components,~\texttt{ASR\_Create\_Dataset} extracts acoustic features from raw waves and generate audio-transcript pairs in the TFRecord format. Users also have the option to enlarge their training sets by various augmentation policies~\cite{DBLP:conf/interspeech/ParkCZCZCL19} by passing optional parameters to this component. In~\texttt{ASR\_Train}, ASR models can be trained from scratch or fine-tuned given training sets, evaluation sets, model configurations and pre-trained model checkpoints (if available) as inputs. EasyASR supports various popular ASR model architectures such as Wav2Letter~\cite{DBLP:journals/corr/CollobertPS16} and Speech Transformer~\cite{DBLP:conf/icassp/ZhaoLWL19}. After training, the component automatically exports the selected checkpoint to the  designated path as a TF SavedModel. The model can be used in the~\texttt{ASR\_Predict} component for fast inference. 

Apart from the three components, EasyASR integrates the technique of extracting wave-transcript pairs from massive video data into the~\texttt{ASR\_Extract\_from\_Video} component to support weakly supervised training of ASR models. Interested readers may refer to~\cite{DBLP:journals/corr/abs-2008-01300} for details. If the performance of the exported model is not satisfactory. Users can call~\texttt{ASR\_Eval} and~\texttt{ASR\_Export} to evaluate and export checkpoints of their own choice, instead of running the fully automatic process provided by~\texttt{ASR\_Train}.

Note that, EasyASR has already equipped its own model zoo trained by our team, containing a number of pre-trained ASR models with high accuracy. Hence, a simple call to~\texttt{ASR\_Predict} can fulfill basic requirements of common ASR applications. The remaining components are designed for domain-specific or other special scenarios.

\noindent\textbf{System Design.}
EasyASR is not a~\textit{machine learning library} but rather a~\textit{machine learning platform for ASR applications}. Hence, various system optimization techniques are designed for large-scale model training. For example, the training procedures in EasyASR are implemented based on the PAISoar framework~\footnote{https://developpaper.com/how-to-improve-the-speed-of-deep-learning-training-100-times-here-comes-paisoar/}, which significantly speeds up the training process distributed across multiple workers and GPUs. The TensorFlow framework we use has been largely optimized to support faster mixed-precision training and have improved communication, memory allocation and I/O mechanisms.

\noindent\textbf{User Interface.}
Despite its sophisticated system and model design, EasyASR is truly EASY to use. On our user interface~\footnote{https://datastudio.dw.alibaba-inc.com/}, only a simple PAI command is required to call the components you need to use.
For example, the following command can be used to i) train an ASR model from scratch (based on the model structure and other settings specified in \texttt{model\_config}) and ii) to export the model to \texttt{model\_export\_dir} for the inference purpose: 
\begin{quote}
\begin{verbatim}
PAI -name ASR_Train 
-Dfinetune=false
-Dconfig='your_path/model_config'
-Dexport='your_path/model_export_dir'
-Dcluster='{"worker": {"count": 4,
    "cpu": 2000, "gpu": 800, 
    "memory": 100000}}';
\end{verbatim}
\end{quote}
Here, four separate workers in the PAI cluster are employed to train the model based on data and model parallelism, each using 20 CPUs, 8 GPUs and 100GB memory.

Specifically, the configuration file~\texttt{model\_config} provides all details on training parameters, settings and the model structure. Take our transformer model as an example, a clip of the configuration file (in JSON) is as follows:
\begin{quote}
\begin{verbatim}
"encoder": TransformerEncoder,
"encoder_params": {
    "encoder_layers": 12, 
    "num_heads": 8,...
  },
"decoder": JointCTCAttenDecoder,
"decoder_params": {
    "attn_decoder": TransformerDecoder,
    "attn_decoder_params": {
      "hidden_layers": 6,
      "num_heads": 8,...
    },
    "ctc_decoder": CTCDecoder,
    "ctc_decoder_params": {...},
  },
"loss": MultiTaskCTCEntropyLoss,
"loss_params": {
    "seq_loss_params": {...},
    "ctc_loss_params": {...},
    "lambda_value": 0.30,
}
\end{verbatim}
\end{quote}
As seen, the model uses both Connectionist Temporal Classification (CTC) and the transformer decoder to generate transcripts. By modifying the configuration file, advanced users have the liberty to customize their models.

\noindent\textbf{Performance.} Based on the improved transformer model described above and the weakly supervised training technique, we have produced state-of-the-art results on Mandarin speech recognition on six public datasets. The experimental results are reported in~\cite{DBLP:journals/corr/abs-2008-01300}.

\section{Related Work and Discussion}

Previously, various deep learning frameworks have been released for training and evaluating ASR models, such as Kaldi~\footnote{https://github.com/kaldi-asr/kaldi}, OpenSeq2Seq~\footnote{https://github.com/NVIDIA/OpenSeq2Seq}, ESPNet~\cite{DBLP:conf/interspeech/WatanabeHKHNUSH18} and wav2letter++~\cite{DBLP:journals/corr/abs-1812-07625}. Our work is different from these frameworks as we integrate our ASR library with the PAI platform for efficient distributed learning. We provide easy-to-use PAI components 
on the platform for users with no re-development needed, and user-customized configurations and modules for advanced developers at the same time.

\section{Conclusion}

In this demo, we present EasyASR, a distributed machine learning platform for learning and serving large-scale, end-to-end ASR models. A simple user interface is created for users to learn ASR models with either pre-defined or user-customized network architectures based on PAI commands. On EasyASR, we produce state-of-the-art results for Mandarin speech recognition. In the future, we will continue to develop our platform to support more state-of-the-art ASR models and make our platform publicly available.

\end{document}